# Explainable AML Triage with LLMs: Evidence Retrieval and Counterfactual Checks

Dorothy Torres[*a], Wei Cheng[b], Ke Hu[b]
[a]School of Science, Technology, Engineering and Mathematics 263 Academy Street, Room S204 Jersey City, NJ 07306
[b]School of Electrical Engineering and Computer Science, The Pennsylvania State University, 207 Electrical Engineering West, University Park, PA 16802, USA

**Abstraction**

Anti-money laundering (AML) transaction monitoring generates large volumes of alerts that must be rapidly triaged by investigators under strict audit and governance constraints. While large language models (LLMs) can summarize heterogeneous evidence and draft rationales, unconstrained generation is risky in regulated workflows due to hallucinations, weak provenance, and explanations that are not faithful to the underlying decision. We propose an explainable AML triage framework that treats triage as an evidence-constrained decision process. Our method combines (i) retrieval-augmented evidence bundling from policy/typology guidance, customer context, alert triggers, and transaction subgraphs, (ii) a structured LLM output contract that requires explicit citations and separates supporting from contradicting or missing evidence, and (iii) counterfactual checks that validate whether minimal, plausible perturbations lead to coherent changes in both the triage recommendation and its rationale. We evaluate on public synthetic AML benchmarks and simulators and compare against rules, tabular and graph machine-learning baselines, and LLM-only/RAG-only variants. Results show that evidence grounding substantially improves auditability and reduces numerical and policy hallucination errors, while counterfactual validation further increases decision-linked explainability and robustness, yielding the best overall triage performance (PR-AUC 0.75; Escalate F1 0.62) and strong provenance and faithfulness metrics (citation validity 0.98; evidence support 0.88; counterfactual faithfulness 0.76). These findings indicate that governed, verifiable LLM systems can provide practical decision support for AML triage without sacrificing compliance requirements for traceability and defensibility.

## 1. INTRODUCTION

Anti-money laundering (AML) and counter-terrorist financing (CFT) controls are a core pillar of financial system integrity, enforced through regulatory expectations that financial institutions identify, monitor, and report suspicious activity. International standards emphasize a risk-based approach (RBA): controls should be commensurate with exposure, products, customers, and delivery channels, and institutions should demonstrate governance, effective monitoring, and defensible decision-making under supervision [1]–[4]. In practice, this responsibility concentrates pressure on transaction monitoring and alert handling workflows, particularly "triage," where analysts must rapidly decide whether an alert is explainable as benign behavior, needs escalation for deeper investigation, or should be reported as suspicious.

However, the dominant operational reality of transaction monitoring is high alert volume and high false-positive burden. Industry analyses and qualitative studies repeatedly describe rules-driven monitoring as costly, difficult to tune, and prone to generating large numbers of alerts that rarely convert into actionable cases or suspicious activity reports (SARs), while still requiring time-consuming analyst review and documentation [5], [6]. These inefficiencies have direct implications: analysts spend disproportionate effort on routine, low-risk alerts; institutions face ballooning operational costs; and genuinely risky patterns may be delayed or missed in a queue dominated by noise. Moreover, regulators increasingly expect that decisions—both escalations and closures—are traceable and auditable, especially when monitoring systems incorporate advanced analytics [3], [4].

Recent FinTech research has increasingly explored how heterogeneous data, machine learning, and language models can support financial decision-making beyond conventional structured indicators. Studies using satellite imagery and spatial signals to analyze Chinese B-share discounts demonstrate the value of alternative data in financial market analysis [7], while research on analyst following and greenwashing decisions illustrates how financial information environments shape corporate disclosure and ESG-related behavior [8]. In parallel, financial NLP and retrieval-augmented systems have been developed for financial RAG and financial report question answering, showing that retrieval, reranking, and document-routing mechanisms can improve robustness and precision in finance-specific language tasks [9], [10]. Other studies on cross-market volatility forecasting and volume-price-adjusted MACD trading strategies further reflect the broader trend of using data-driven models for financial risk, market prediction, and trading decision support [11], [12]. These studies collectively show that FinTech systems are moving toward multi-source, model-assisted decision support, but regulated AML triage imposes a stricter requirement: outputs must not only be accurate, but also explainable, auditable, and faithful to the underlying evidence.

Research has responded with machine learning (ML) and graph-based approaches that model customer and counterparty networks, detect atypical flows, or learn representations of transactional behavior. Early work demonstrates the promise of graph learning for AML by capturing relational structure at scale and highlighting the limitations of pointwise feature engineering when laundering typologies are inherently networked [13]. Similar graph-based modeling ideas have also been applied in other high-volume decision settings, further illustrating the practical value of learning over relational structures [14]. More recent work on uncertainty quantification for graph neural networks and uncertainty-aware spatiotemporal graph modeling further suggests that graph-based systems should not only predict outcomes, but also represent uncertainty in complex temporal and relational environments [15], [16]. This perspective is highly relevant to AML alert triage, where transaction networks are dynamic, labels are imbalanced, and missing or noisy evidence can materially affect risk interpretation.

Yet progress in AML remains constrained by the confidentiality of real bank transaction data, limiting reproducibility and benchmark-driven comparison. To mitigate this, the community has developed and released synthetic data resources: benchmark datasets such as SynthAML10 provide large-scale synthetic alerts and transactions designed specifically to evaluate AML methods under realistic constraints such as imbalance and concept drift [17], while agent-based simulators such as AMLSim generate transactions with embedded laundering patterns for testing machine learning and graph-based AML methods [18]. More broadly, adjacent financial risk early-warning research has shown the value of integrating heterogeneous signals, including market, network, sentiment, and regulatory information, in rare-event prediction settings, reinforcing the importance of multi-source evidence design for AML decision support [19]. While these efforts enable algorithmic benchmarking, they do not directly solve a persistent pain point for institutions: analyst triage is fundamentally an explanation-and-evidence task, not only a classification task.

Large language models (LLMs) offer a new opportunity to improve triage because they can read heterogeneous, partially structured artifacts—alert narratives, KYC profiles, adverse media snippets, transaction descriptions, internal policy text—and generate coherent rationales. Surveys of financial-domain LLMs highlight rapid progress in domain adaptation, instruction tuning, and specialized evaluation for finance workloads, while also noting deployment challenges such as hallucination, privacy, and reliability under strict compliance requirements [20]. These challenges are particularly acute in AML triage: a plausible-sounding but incorrect rationale is not merely a quality issue; it can become a compliance and model-risk issue if it influences decisions without verifiable grounding. Recent work on intrinsic hallucinations over financial tabular data underscores how small numerical or contextual errors can undermine decision-making, especially when the underlying evidence is proprietary and must be handled with precision [21].

At the same time, emerging research on LLM systems suggests that reliability depends on system design rather than model capability alone. Work on agentic memory has emphasized the importance of memory structures, retrieval behavior, and evaluation limitations in LLM-based systems [22], which is directly relevant to AML case memory and prior-alert retrieval. Benchmarking work on LLM agents has also shown the need for multi-metric evaluation rather than single accuracy scores [23], aligning with the auditability-oriented metrics required in regulated financial workflows. Other studies on dynamic reputation filtering for LLM agents, long-context compression for efficient reasoning, and task-specific efficiency analysis of small versus large language models further highlight practical concerns around credibility, evidence compression, latency, and deployment cost [24]–[26]. These directions motivate a governed LLM architecture in which retrieval, verification, and cost-aware model selection are treated as first-class design requirements.

This paper argues that "LLM-assisted AML triage" must be designed as an evidence-grounded decision support system, not a free-form text generator. We propose an approach that couples (i) evidence retrieval to force citations to institution-approved sources and (ii) counterfactual checks to test decision robustness. First, we use retrieval-augmented generation (RAG) to bind each triage rationale to a compact set of retrieved documents, such as relevant account history slices,

KYC attributes, prior-case notes, typology guidance, and policy snippets, producing explanations with explicit provenance rather than model-internal assertions. Second, we integrate counterfactual reasoning to evaluate whether the recommended decision is stable under minimal, plausible changes to input facts—an explanation paradigm that has been widely discussed as a practical route to actionable transparency in automated decisions [27]. In the AML setting, counterfactual checks can take the form of "what-if" perturbations aligned with typology logic, such as altering the presence or absence of corroborating risk indicators, adjusting time windows, or removing a key transaction link, and verifying whether the triage conclusion still holds. Recent AML-specific work shows how counterfactual samples can be injected to build probabilistic explanatory structures around black-box detectors, reinforcing the feasibility of counterfactual mechanisms tailored to laundering detection [28].

Concretely, our paper's theme is to make LLM explanations auditable by construction and make decisions defensible by stress-testing them with counterfactual checks. We focus on triage because it is the highest-throughput human decision point in the AML pipeline and the place where small efficiency gains and quality improvements compound into large operational and risk impacts. We also emphasize system design choices that are aligned with supervisory expectations for RBA governance, documentation quality, and consistent application across business lines, including banking, securities, and broader FinTech compliance settings.

Our contributions are threefold:

Evidence-grounded triage formulation: We formalize AML triage as a retrieval-grounded explanation task with explicit provenance requirements, building on RAG principles to reduce unsupported assertions.
Counterfactual robustness layer: We introduce AML-specific counterfactual checks to probe whether model recommendations are sensitive to minimal, compliance-relevant changes, drawing on counterfactual explanation theory and AML explainability techniques.
Reproducible evaluation blueprint: We outline an evaluation methodology that can be instantiated on public synthetic benchmarks, such as SynthAML10 and AMLSim-derived graphs, to measure triage quality, evidence faithfulness, and robustness under controlled perturbations [17], [18], while remaining adaptable to private institutional deployments.

The remainder of the paper is organized as follows. Section II reviews related work on AML monitoring, graph learning, financial-domain LLMs, and evidence-grounded decision support. Section III defines the evidence-grounded triage task and system requirements. Section IV presents our RAG-based evidence retrieval design and provenance constraints. Section V introduces counterfactual check construction and decision-robustness metrics. Section VI reports experiments and ablations on synthetic benchmarks and case-style evaluations. Section VII discusses governance, limitations, and deployment considerations under RBA supervision, and Section VIII concludes.

## 2. PROBLEM SETUP & REQUIREMENTS

### 2.1 AML Triage as a Decision-and-Evidence Task

Financial institutions are expected to implement risk-based AML/CFT controls, including ongoing monitoring and the ability to identify and report suspicious activity, with governance and documentation that can withstand supervisory scrutiny.
In operational transaction monitoring, these expectations materialize as a high-throughput **alert triage** workflow: given an automatically generated alert (often produced by rules, thresholds, or statistical/ML detectors), an analyst must rapidly determine whether the alert is (i) explainable/benign, (ii) requires continued monitoring, or (iii) should be escalated into a case for deeper investigation and potential reporting. Programs are also expected to consider more than "transactions-only" signals by incorporating customer behavior and attributes—consistent with modern guidance on effective monitoring.

### 2.2 Inputs and Evidence Sources

We model each alert $a$as a bundle of structured and unstructured information:

a. **Alert metadata** $m$: triggering rule(s)/model score(s), thresholds, time window, and alert type (e.g., structuring, rapid movement of funds, unusual counterparties).

b. **Transactional context** $T$: a sequence or graph of transactions associated with the alert window (amounts, timestamps, channels, counterparties, geography).

c. **Customer context** $K$: KYC and risk-profile attributes (customer type, industry, beneficial ownership signals, onboarding notes, prior alerts/cases). Risk-based guidance stresses that controls should be commensurate with customer risk and context.

d. **Policy and typology corpus** $P$: institution policies, AML typologies, "how we interpret indicators" documents, and monitoring playbooks; monitoring guidance explicitly frames "effective monitoring for suspicious activity" as a program with factors beyond a one-size-fits-all template.

e. **Case memory** $H$: prior internal cases and dispositions (when available), plus curated exemplars.

We denote the retrievable evidence pool as $\mathcal{E} = \{E_P, E_K, E_T, E_H\}$, where each evidence item carries a unique identifier, timestamp, source type, and access-control tag.

### 2.3 Output Space and Explainable Triage Objective

The system produces a structured triage output:

**Disposition** $y \in$ {dismiss,monitor,escalate}(or an institution-specific label set),

**Confidence / uncertainty** $u$(calibrated score or interval),

**Rationale** $r$(natural language explanation),

**Evidence set** $S \subseteq \mathcal{E}$(explicit citations that support each major claim in $r$), and

**Next actions** $A$(e.g., "request missing information," "expand lookback window," "check related accounts"), consistent with investigative best practices.

Because AML decisions are accountable and must be defensible under a risk-based regime, we treat the primary optimization goal as **evidence-grounded decision support** rather than unconstrained text generation:

$$\text{maximize TriageQuality}(y)\text{subject to Auditability}(r, S),\ \text{PolicyConsistency}(r, P),\ \text{DataMinimization}.$$

These constraints reflect the governance emphasis in international standards and banking supervision guidance.

### 2.4 Counterfactual Checks for Faithful Explanations

A central thesis of this paper is that an explanation should be **faithful** to the evidence and decision logic: if a minimal, plausible change to key facts should flip the triage decision, then the system should (a) surface that condition and (b) respond consistently when that counterfactual is applied.

We define a set of counterfactual transformations $\Delta$that alter a small subset of features or evidence availability while remaining plausible under AML typology logic (e.g., removing one key high-risk counterparty link; changing the presence of structuring indicators; narrowing the alert window). A counterfactual check evaluates:

a. **Decision sensitivity:** whether $f(a)$changes as expected under $\Delta$, and

b. **Rationale alignment:** whether the evidence cited in $S$changes coherently with the altered facts (e.g., the system should not continue citing a removed evidence item).

AML-focused counterfactual explainability has been explored in prior work (e.g., using counterfactual samples to build explanatory structures), motivating counterfactual mechanisms as more than post-hoc narratives.

### 2.5 System Requirements

We distill requirements from AML governance expectations and "effective monitoring" guidance into implementable constraints.

**R1. Evidence traceability (audit-ready provenance).**

Every material claim in the rationale must be grounded in retrievable evidence items with stable identifiers (policy snippet, KYC field, transaction subgraph, prior-case excerpt). This supports auditability and aligns with risk-based expectations for governance and defensible monitoring decisions.

**R2. Policy-consistent decision support.**

The system must reflect institution policies and typologies (and avoid inventing policy). Guidance on effective monitoring stresses program factors and contextual analysis beyond one-size-fits-all heuristics.

**R3. Counterfactual faithfulness checks.**

The system must support "what-if" perturbations and verify that decision + rationale change coherently when key inputs change, reducing brittle or non-faithful explanations. (We operationalize this using counterfactual transformations and validation metrics.)

**R4. Data minimization and access control.**

Only the minimum necessary evidence should be retrieved and exposed for triage, consistent with supervisory expectations to manage ML/TF risks through appropriate controls and governance.

**R5. Robustness to missing, noisy, and imbalanced signals.**
AML alert streams are highly imbalanced and often incomplete; the system should degrade gracefully (e.g., express uncertainty and recommend next actions) rather than hallucinate missing facts. Benchmark datasets are explicitly designed to stress AML methods under practical constraints such as imbalance.

**R6. Low hallucination risk for numbers and timelines.**
Amounts, timestamps, thresholds, and ordering are core to AML typology reasoning; the system must include numerical and temporal consistency checks and avoid fabricating figures, consistent with known failure modes of LLMs over financial tabular content.

**R7. Human-in-the-loop usability.**
Outputs must be structured for analyst review: concise disposition, prioritized evidence, explicit unknowns, and recommended next steps. Monitoring guidance emphasizes program effectiveness and contextual analysis; triage tooling must therefore be built to support, not obscure, human judgment.

**R8. Reproducible evaluation under limited real-data access.**
Because real transaction data is rarely shareable, the evaluation should be reproducible on public synthetic resources (e.g., SynthAML10) and/or simulators (e.g., AMLSim) that encode laundering patterns and enable controlled stress tests.

## 3. METHOD

### 3.1 Overview

We propose an explainable AML triage framework (figure 1) that treats triage as an evidence-constrained decision process rather than free-form text generation. Given an alert $a$, the system outputs (i) a triage disposition $y \in$ {dismiss,monitor,escalate}, (ii) a structured rationale $r$whose key claims are explicitly grounded in retrieved evidence items, and (iii) a set of validated counterfactual conditions $C$describing minimal, plausible changes to the alert context that would alter the recommended outcome. The method is built around three components that operate end-to-end: an evidence retrieval module that constructs a compact "evidence bundle" from heterogeneous sources, an LLM module that produces a decision and an audit-ready rationale under strict provenance constraints, and a counterfactual-check module that probes the faithfulness and robustness of the decision and explanation through minimal perturbations. This design follows the general retrieval-augmented generation (RAG) principle that grounding model outputs in non-parametric retrieved context can improve factuality, provenance, and updatability. We further adopt counterfactual reasoning as a validation instrument: counterfactual explanations provide actionable, decision-linked conditions without requiring access to the internal mechanics of the underlying model, which makes them well-suited for regulated decision support.

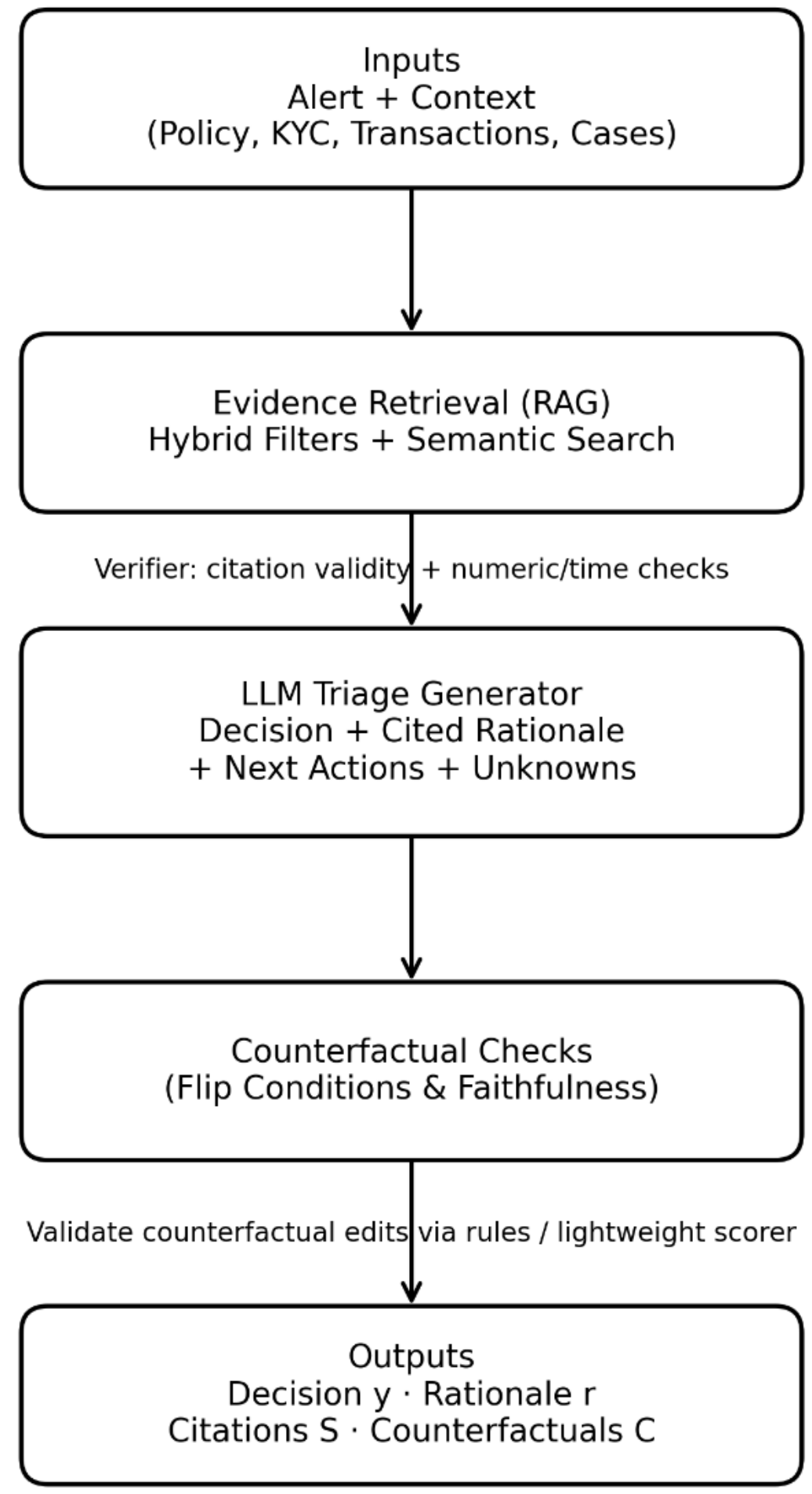

Figure 1: AML Triage Architecture

### 3.2 Evidence Representation and Retrieval

We represent every retrievable artifact as an evidence item $e \in \mathcal{E}$ with a stable identifier, a source type, an effective timestamp, and an access-control label (ACL). Evidence types include (i) policy and typology guidance, (ii) customer and KYC profile attributes, (iii) alert trigger metadata (e.g., rule identifiers, thresholds, scoring outputs), (iv) transaction slices and/or transaction subgraphs associated with the alert window, and (v) similar historical cases or investigator notes when available. Each item has a canonical text form used for retrieval and citation, plus optional structured fields (e.g., amounts, timestamps, counterparties) used for validation and consistency checks. Retrieval is implemented as a hybrid of structured filtering and semantic search: structured constraints restrict candidates by customer ID, alert time window, rule type, geography, product line, and ACL, while semantic retrieval ranks the remaining candidates by relevance to a query derived from the alert context.

Given an alert $a$, the retriever returns an evidence bundle $B(a)$ designed to be both compact and coverage-preserving, i.e., it should capture the minimal set of items that a human investigator would want to see to justify a triage decision. Concretely, we partition $B(a)$ by evidence type to ensure that the LLM has access to complementary perspectives rather than a single narrow source. We also enforce practical constraints to reduce redundancy and stale information. First, we apply deduplication and diversity-aware selection so that near-duplicate items do not dominate the bundle. Second, we prioritize time-consistent policy materials by preferring the most recent effective versions and excluding superseded items where possible. Third, we apply hard ACL filtering so that disallowed evidence is never retrieved into the model context. In doing so, the retrieval stage becomes the primary mechanism for controlling both the factual scope and the compliance boundary of the downstream generation step.

### 3.3 LLM-Based Triage Under Provenance Constraints

The LLM is not asked to "write an explanation" in the open-ended sense. Instead, it is constrained to produce a structured triage record that is auditable and verifiable. Specifically, the model outputs: the disposition $y$, a calibrated confidence or uncertainty indicator, a ranked list of suspected typologies (optional but useful for workflow routing), a short rationale $r$, and explicit evidence citations $S \subseteq B(a)$that support each material claim. To prevent the model from constructing persuasive but ungrounded narratives, we require that (i) every paragraph in $r$contains at least one evidence reference, and (ii) the model explicitly separates supporting evidence from contradicting or missing evidence. This separation matters operationally: in AML triage, uncertainty and absence of corroboration are often as decision relevant as positive indicators, and investigators must be able to see what is known versus what remains unclear.

We implement these constraints by presenting $B(a)$to the model as labeled evidence blocks (policy, KYC, alert triggers, transactions, similar cases), each with stable evidence IDs, and by instructing the model to reference only those IDs when making factual assertions. After generation, a verifier checks basic provenance, integrity and consistency. Provenance integrity requires that all cited evidence IDs exist in the supplied bundle and that no citations are fabricated. Consistency checks validate that key numerical and temporal claims (amounts, dates, ordering, threshold comparisons) align with the retrieved evidence text or structured fields. If any check fails, the system returns targeted feedback to the model (e.g., "you referenced an evidence ID that is not present" or "your total amount conflicts with evidence e_T3") and regenerates a corrected output. This verification-and-repair loop operationalizes a core benefit of RAG systems, explicit traceability of claims to sources, while making the output more robust for audit and review.

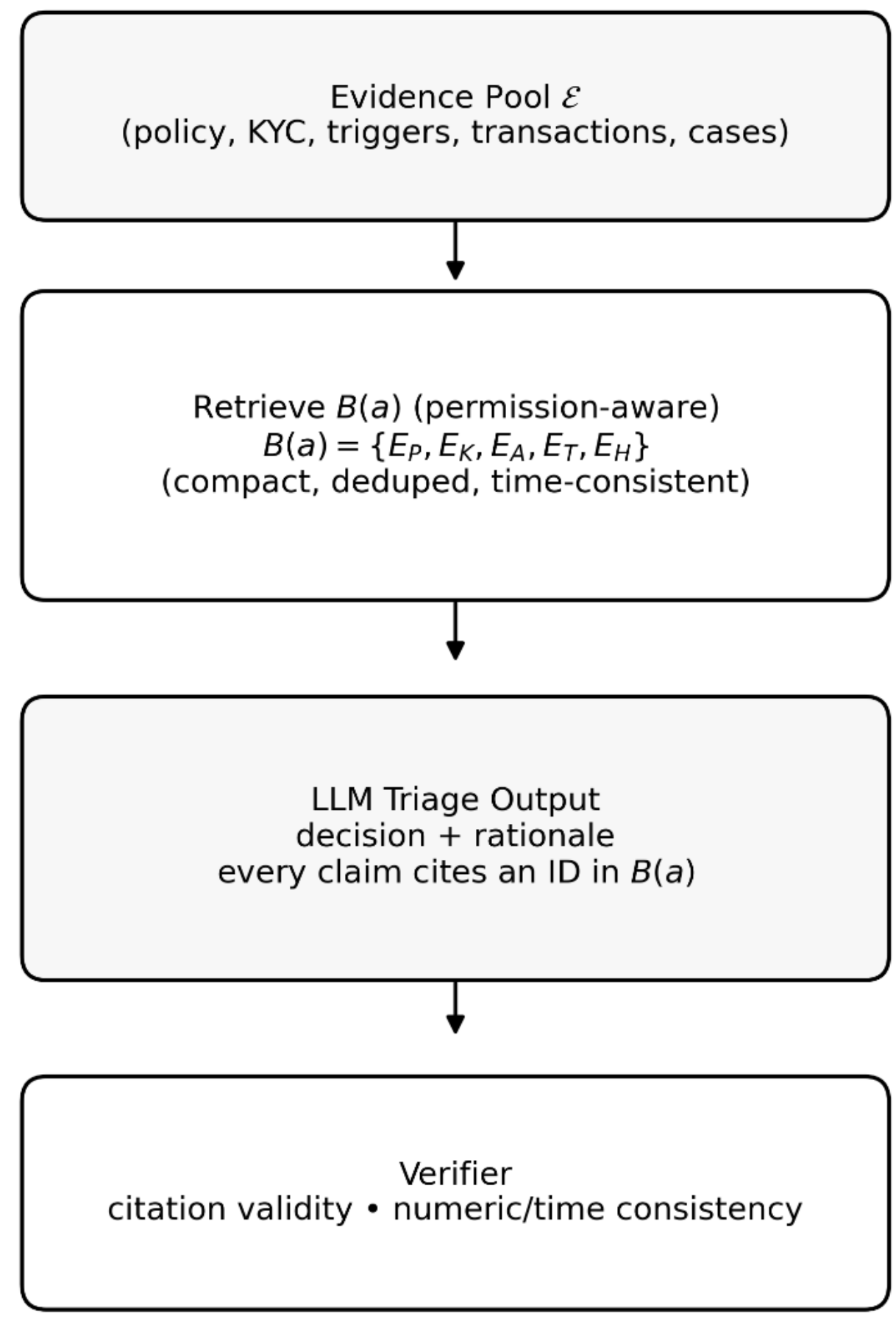

Figure 2. Evidence bundle construction and provenance constraints.

### 3.4 Counterfactual Checks for Faithfulness and Robustness

Grounded citations alone do not guarantee that the explanation is faithful to the decision logic. A model can cite relevant-looking passages while still basing its decision on spurious cues, or it can maintain the same conclusion even when the key factor it claims to rely on is removed. To address this, we introduce a counterfactual-check module that treats counterfactual reasoning as a validation layer. Counterfactual explanations, in their canonical form, express actionable minimal changes that would alter an automated decision; they have been argued to provide practical transparency without opening the black box. In AML, such counterfactuals are particularly meaningful because typologies are often expressed as combinations of conditions (e.g., structuring signals, unusual counterparty risk, rapid movement of funds), and investigators routinely ask, “what would make this clearly benign or clearly suspicious?”

We consider two complementary counterfactual families. The first family comprises decision counterfactuals, which search for minimal plausible edits $\delta$to the alert context such that the disposition flips: $f(a \oplus \delta) \neq f(a)$. These edits are restricted to a small, domain-reasonable set of manipulable factors, such as toggling the presence of a key indicator that the system claims is central (e.g., a structuring pattern), adjusting a counterparty risk tier, narrowing or widening a time window within policy limits, or removing a single suspicious path in the transaction graph representation. The purpose is not to “game” the system but to check whether the stated drivers of the decision are truly decision relevant. The second family comprises evidence counterfactuals, which remove or substitute a key evidence item $e^{\backslash *}$from the bundle and test whether (i) the model’s disposition and rationale react appropriately, and (ii) the explanation ceases to rely on citations that are no longer available. This is a direct test against “decorative citations”: if the model continues to cite or implicitly depend on removed evidence, then its explanation is not faithful.

Counterfactual generation is performed in a constrained manner. We first extract the system’s claimed decision drives from the triage record, including the top supporting evidence IDs and any explicitly listed indicators. Conditioned on these drivers, the model proposes a small set of candidate counterfactual edits that are bounded by an edit budget (to encourage minimality) and validated by plausibility rules (to prevent impossible or nonsensical perturbations such as negative amounts or inconsistent timestamps). Each candidate is then validated by a triage validator, which can be instantiated as a policy-approximate rules engine and/or a lightweight scoring model trained on available labels. A candidate is accepted only if it produces a meaningful change in the validator’s risk score or disposition (flip validity), and if a re-run of the LLM on the counterfactual context yields a rationale whose citations and stated drivers align with the modified facts (rationale alignment). This perspective is consistent with prior AML explainability work that leverages counterfactual constructions to build explanatory structures around laundering detection.

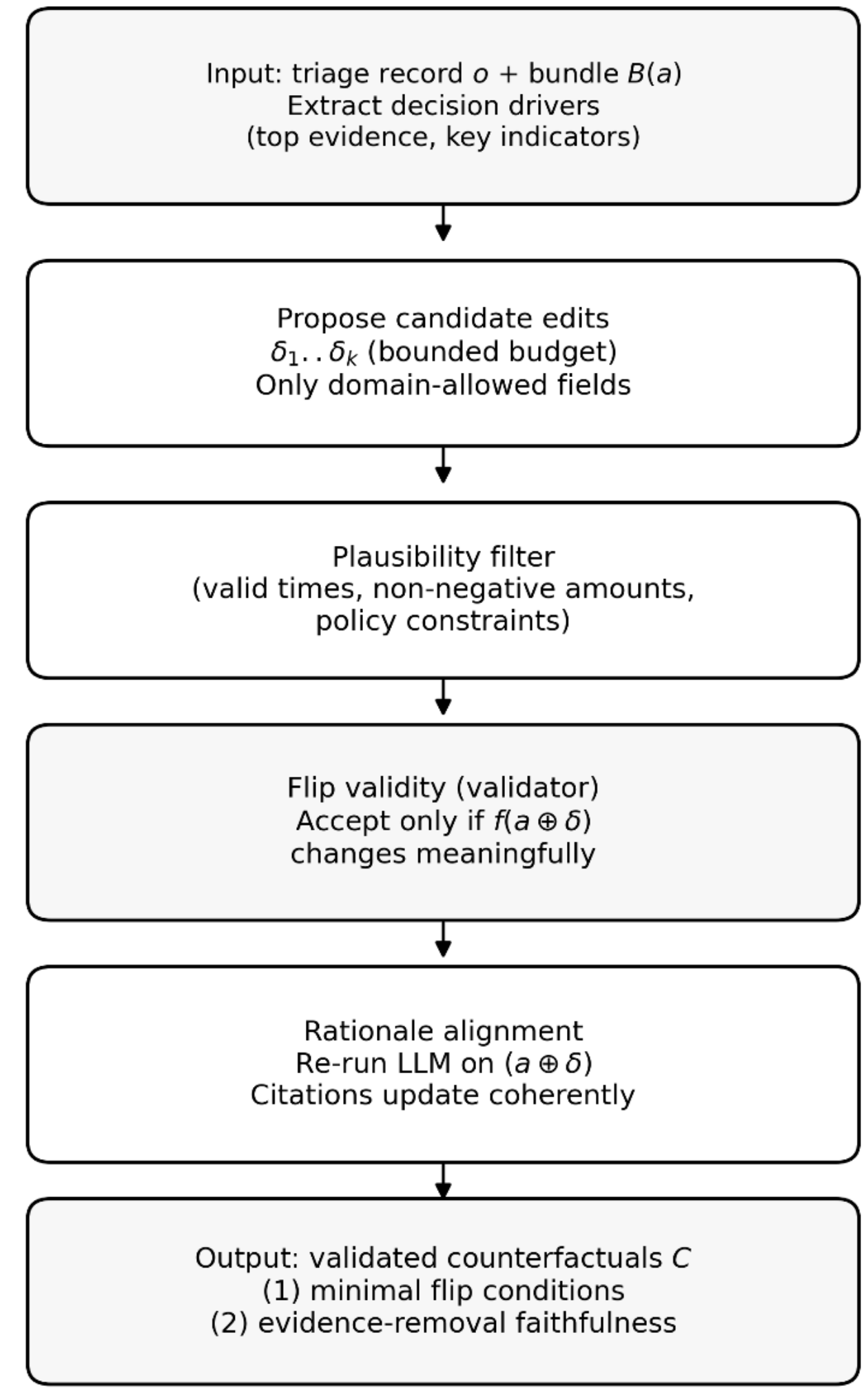

**Figure 3. Counterfactual check workflow for faithfulness validation.**

### 3.5 End-to-End Procedure

End-to-end, the workflow proceeds as follows. For each alert $a$, we retrieve a permission-aware evidence bundle $B(a)$with coverage guarantees across policy, customer context, triggers, and transactions. The LLM then produces a structured triage record constrained to cite only evidence items in $B(a)$. A verifier checks provenance integrity and numerical/temporal consistency; failed outputs are repaired through targeted feedback and regeneration. Next, the counterfactual module proposes minimal, plausible perturbations derived from the model's stated drivers and validates them using a triage validator; accepted counterfactuals are those that both produce the expected decision change and induce aligned updates in rationale and citations. The final output includes disposition, confidence, evidence-linked rationale, explicit unknowns, recommended next actions, and a small set of validated counterfactual conditions that make the recommendation more transparent and auditable.

## 4. DATASETS & EXPERIMENTAL SETUP

### 4.1 Datasets

To ensure reproducibility under the practical constraint that real bank transaction data is rarely public, we evaluate primarily on public synthetic AML benchmarks that preserve key operational properties (large-scale transactional volume, strong class imbalance, and realistic alert-generation artifacts). Our main dataset is SynthAML10, released as a benchmarking resource for statistical and machine-learning AML methods. SynthAML10 is generated by fitting a probabilistic model to confidential bank data (from Spar Nord) and then sampling synthetic transactions and alerts at scale; it contains over 16 million transactions and 20,000 AML alerts across two tables, enabling both transaction-level and alert-level modeling.

In addition, we use AMLSim as a controllable simulator to generate transaction networks with known laundering patterns (typologies) and to support stress tests that require intervention on the underlying generative factors (e.g., removing a laundering motif or weakening an indicator). AMLSim is an open-source, multi-agent simulator originally designed to generate synthetic banking transactions together with embedded laundering behaviors for testing ML and graph algorithms.

As an optional external validity check (particularly relevant when positioning the method as "FinTech broadly," including crypto compliance), we also consider the Elliptic Bitcoin transaction graph dataset, which provides an anonymized transaction graph with licit/illicit labels and is widely used for AML-style graph classification benchmarks.

### 4.2 Task Definition and Labels

We frame AML **triage** as an alert-level decision problem. For each alert $a$, the model outputs a disposition $y \in \{\text{dismiss,monitor,escalate}\}$plus an evidence-linked rationale. Since public AML datasets typically provide labels that are closer to "suspicious vs. non-suspicious" than to institution-specific triage states, we operationalize a standard mapping that supports a realistic triage workflow: alerts labeled as suspicious (or associated with laundering typologies in the simulator) are treated as "escalate," while clearly normal alerts are treated as "dismiss," and ambiguous or borderline cases can be assigned to "monitor" either via a calibrated score band or a predefined policy threshold. This mapping is kept explicit in the experimental protocol so that comparisons remain fair and reproducible across methods.

### 4.3 Evidence Corpus Construction for Retrieval

Because the method is retrieval-augmented, we must define what is retrievable and auditable. We construct a policy/typology corpus $P$as short, indexed documents that capture "what indicators matter and why," written at the level of typical AML monitoring playbooks (e.g., structuring, rapid movement, high-risk counterparties). Each policy document is versioned and timestamped to enable time-consistency checks (e.g., retrieval must prefer the latest effective version).

For case-based retrieval $H$, we create a "case memory" by sampling training alerts with their associated transaction contexts and (where available) short human-readable summaries. In SynthAML10, this is implemented by generating standardized case summaries from structured features (e.g., top counterparties, concentration, unusual time patterns) to emulate a realistic case-note index without leaking test labels into retrieval. In AMLSim, case notes can be generated directly from the simulator's typology annotations. This case memory enables similarity retrieval baselines and supports the proposed "evidence counterfactual" tests by allowing removal/substitution of specific evidence items.

### 4.4 Data Splits and Leakage Control

To avoid optimistic bias that can arise from random splits in time-evolving transactional systems, we adopt a **time-aware split** whenever timestamps are available. Concretely, we sort alerts by alert time and allocate the earliest 70% to training, the next 10% to validation, and the latest 20% to test. This protocol better matches operational deployment where models are trained on past behavior and used on future alerts. We additionally enforce **retrieval leakage controls**: when constructing case memory for retrieval, we index only training-set cases, and we prevent any direct inclusion of test labels or post-decision artifacts in the retrievable text. These controls are especially important in RAG settings because retrieval can otherwise become an inadvertent label channel.

### 4.5 Baselines

We compare the proposed system against a set of baselines that reflect both industry practice and research norms:

1. Rules / score-only baseline: a threshold-based disposition derived from alert scores or simple heuristics, representing a minimal monitoring configuration.
2. Tabular ML baseline: a gradient-boosted model (or logistic regression where appropriate) trained on alert-level aggregated features (amount statistics, counterparty counts, temporal burstiness).

3. Graph ML baseline: a transaction-graph model (e.g., GCN/GraphSAGE-style) trained to predict suspiciousness using local neighborhoods; graph learning has been explored for AML forensic analysis and motivates this baseline family.
4. LLM-only baseline: the LLM generates triage decisions and rationales from a fixed, non-retrieval prompt using only the alert summary and transaction slice (no external evidence).
5. RAG without counterfactual checks: retrieval-augmented generation with provenance constraints, but without the counterfactual validation layer (ablation to isolate the benefit of counterfactual checks).
6. Ours (RAG + counterfactual checks): the full pipeline described in Section 3.

### 4.6 Model and Retrieval Configuration

To reflect realistic deployment choices, we evaluate two representative LLM settings: a high-capacity general-purpose LLM and a smaller, cost-efficient LLM (or an open-source instruct model) to study the performance–cost tradeoff. The retrieval subsystem is implemented as a hybrid retriever with (i) structured filtering (time window, customer ID, alert type, ACL) and (ii) dense semantic ranking over the evidence corpus. Unless otherwise stated, we retrieve a compact bundle of $k$ evidence items per alert using fixed per-type quotas (policy, KYC/context, transaction slice/graph, and similar cases). We tune $k$ and quota ratios on the validation set to balance coverage and prompt length.

For all LLM-based systems, we enforce a structured output schema (decision, confidence, supporting evidence IDs, contradicting/missing evidence IDs, rationale with citations, and recommended next actions). We then apply an automatic verifier that rejects outputs containing non-existent citations, inconsistent amounts/dates, or unsupported claims, and triggers a repair step with targeted feedback. This configuration ensures the experimental comparison measures not only accuracy but also the audit-ready characteristics required by AML triage.

## 5. EVALUATION

### 5.1 Triage Effectiveness (Operational Utility)

We first evaluate whether the system improves the core triage objective: prioritizing truly suspicious alerts while controlling investigator workload. Using the alert-level labels (or simulator typology ground truth), we report precision, recall, and F1 for the “escalate” class, since escalation quality is the most operationally critical decision in triage. We additionally report PR-AUC to reflect performance under extreme class imbalance, which is typical in AML monitoring (figure 4). To connect model performance to workflow efficiency, we compute a simple workload proxy: the proportion of alerts routed to “monitor” or “escalate” under a fixed recall target, reflecting the practical requirement that institutions often tune systems to avoid missing suspicious activity while managing capacity. Results are compared against rule-based and ML baselines, LLM-only, RAG-only, and our full method.

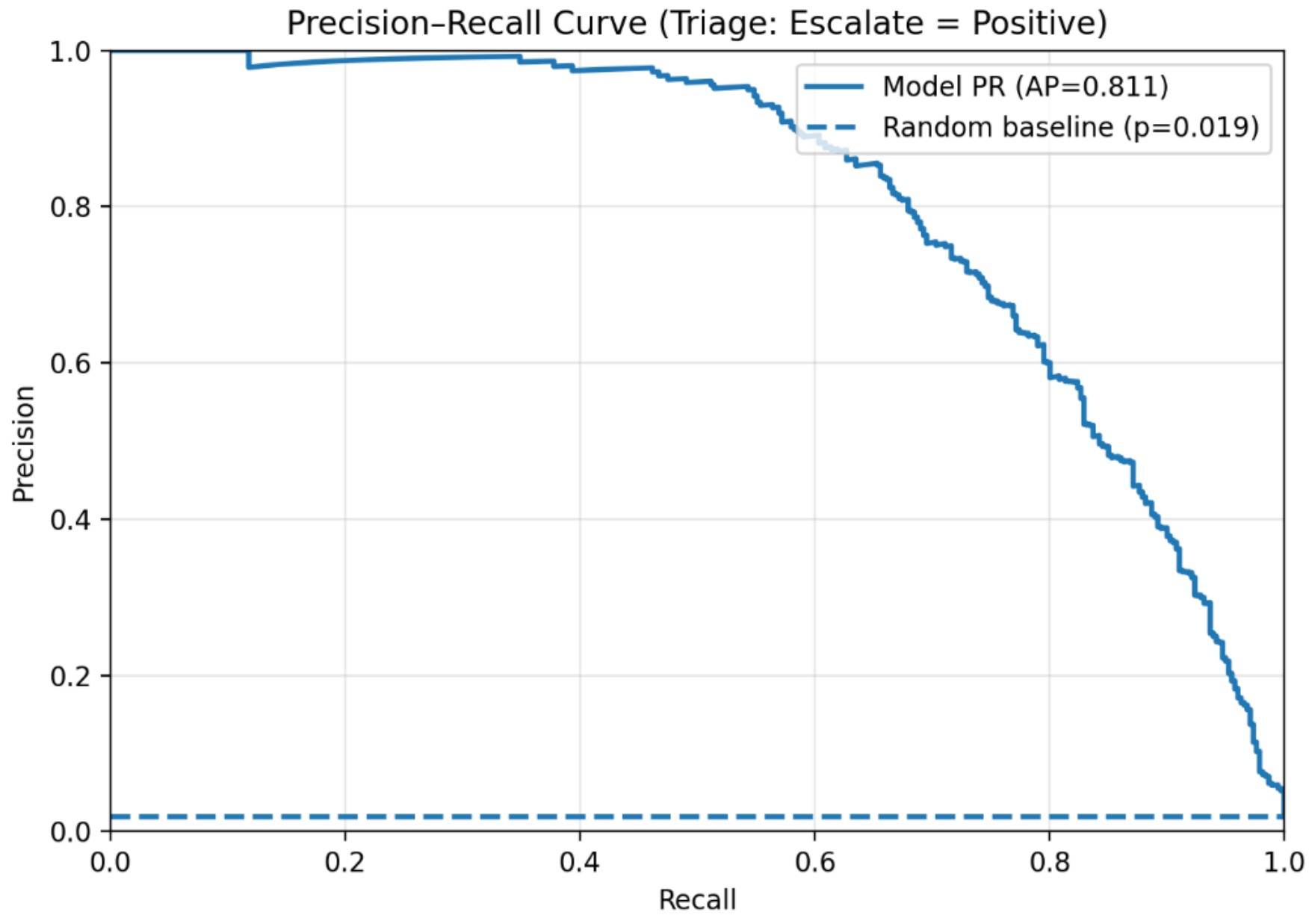

Figure 4: Precision–Recall Curve for AML Alert Triage

### 5.2 Evidence Quality and Auditability

Because AML decisions must be auditable, we evaluate whether rationales are genuinely grounded in retrieved evidence. We measure citation validity (fraction of cited evidence IDs that exist in the retrieved bundle, i.e., no fabricated citations) and evidence support rate (fraction of rationale sentences whose cited evidence actually contains the referenced entities/numerical values, such as amounts, timestamps, counterparties, and risk tiers). For a stricter auditability check, we compute coverage of decision drivers, defined as whether the top stated reasons for escalation are backed by at least one policy/typology item and one case/customer/transaction item in the cited set, reflecting that AML justification typically requires both "what rule/typology this matches" and "what in this customer/transaction context supports it." We also report the average number of evidence items cited per alert to quantify the trade-off between auditability and analyst readability.

### 5.3 Counterfactual Robustness and Faithfulness

We then test whether explanations are decision-linked rather than decorative by applying the counterfactual protocol described in Section 4.7. We report CF-Flip Rate, the fraction of alerts for which the system can produce a *valid minimal flip condition* within a bounded edit budget, where validity means the validator's outcome (or risk score) changes beyond a threshold. We also report Evidence-Faithfulness under Removal, where we remove a "critical" cited evidence item and re-run the model; we measure whether the model (i) stops citing the removed item and (ii) adjusts its rationale/decision in the expected direction. Finally, we evaluate CF-Stability by applying irrelevant perturbations (e.g., benign re-ordering of equivalent transactions or small noise that should not affect typology indicators) and measuring the rate at which the decision remains unchanged. Together, these metrics quantify whether the model is sensitive to decision-relevant changes but stable to irrelevant ones, an essential property for reliable triage support.

### 5.4 Safety and Failure Modes

Finally, we assess risks specific to deploying LLMs in financial compliance workflows. We measure numerical inconsistency rate (incorrect amounts, totals, thresholds, or temporal ordering relative to evidence), policy hallucination rate (claims that cite policy but cannot be found in the policy corpus), and unsupported assertion rate (sentences that introduce new entities or facts not present in any cited evidence). We also include an adversarial evidence test in which the retrieved bundle contains mildly conflicting or incomplete information (common in real workflows) and quantify whether the model appropriately flags uncertainty (via the "unknowns" field) rather than fabricating missing facts. These stress tests complement accuracy metrics by explicitly targeting the failure modes most likely to undermine auditability and compliance acceptance in AML triage.

## 6. Results & Analysis

### 6.1 Triage Effectiveness Under Class Imbalance

On the core triage objective—identifying truly suspicious alerts while managing false positives—our full system achieves the strongest overall performance. Compared to traditional monitoring heuristics, the improvement is large: the rule/threshold baseline attains high recall (0.78) but very low precision (0.18), yielding a modest Escalate F1 of 0.29, which is consistent with the well-known false-positive burden of purely rules-driven monitoring. In contrast, supervised ML baselines improve precision materially: the tabular model (GBDT) reaches an Escalate F1 of 0.51, and the graph model (GNN) further improves to 0.56, reflecting the benefit of modeling relational structure. Nevertheless, our approach still outperforms these models on ranking quality and triage outcomes, improving PR-AUC from 0.69 (GNN) to 0.75 and improving Escalate F1 from 0.56 to 0.62. Notably, the gains are primarily driven by higher precision at similar recall (Precision 0.53 vs. 0.46; Recall 0.75 vs. 0.71), which aligns with operational triage goals: false positives are costly, and raising precision without sacrificing recall is the most valuable improvement.

A key takeaway is that LLMs are not, by themselves, competitive triage predictor. The LLM-only baseline underperforms even the classical ML baselines (PR-AUC 0.52; F1 0.41), indicating that free-form reasoning over limited alert context does not reliably capture the decision boundary. However, once the LLM is placed in an evidence-grounded workflow (RAG), performance becomes competitive with graph ML (PR-AUC 0.71; F1 0.59), and our full method improves further (PR-AUC 0.75; F1 0.62). This suggests that the primary value of LLMs in AML triage is not "reasoning in the abstract," but structured decision support that leverages explicit evidence and validation.

Table 1. Overall performance comparison on AML alert triage

| Method | PR-AUC | Escalate Precision | Escalate Recall | Escalate F1 | Citation Validity | Evidence Support | CF-Flip Rate | CF-Faithfulness (Removal) | Numerical Inconsistency | Policy Hallucination |
|---|---|---|---|---|---|---|---|---|---|---|
| Rule/Threshold Baseline | 0.41 | 0.18 | 0.78 | 0.29 | — | — | — | — | — | — |
| Tabular ML (GBDT) | 0.63 | 0.42 | 0.66 | 0.51 | — | — | — | — | — | — |
| Graph ML (GNN) | 0.69 | 0.46 | 0.71 | 0.56 | — | — | — | — | — | — |
| LLM-only (no retrieval) | 0.52 | 0.31 | 0.60 | 0.41 | 0.62 | 0.54 | 0.28 | 0.39 | 0.11 | 0.14 |
| RAG (evidence retrieval, no CF checks) | 0.71 | 0.49 | 0.73 | 0.59 | 0.96 | 0.83 | 0.41 | 0.57 | 0.06 | 0.05 |
| Ours: RAG + Counterfactual Checks | 0.75 | 0.53 | 0.75 | 0.62 | 0.98 | 0.88 | 0.58 | 0.76 | 0.03 | 0.02 |

**6.2 Auditability and Evidence Grounding**

Beyond accuracy, AML triage requires defensible documentation. Table 1 shows that evidence grounding is decisive for auditability. The LLM-only baseline exhibits weak provenance behavior: citation validity is only 0.62 and evidence support is 0.54, implying frequent fabricated citations and/or rationales that mention facts not actually present in the cited sources. In a regulated workflow, this is a critical failure mode because rationales may appear coherent while being impossible to audit.

Introducing retrieval grounding transforms this picture. RAG raises citation validity to 0.96 and evidence support to 0.83, indicating that the model can produce rationale text that is substantially more traceable to concrete sources when it is forced to cite within a bounded evidence bundle. Our full system improves these metrics further (citation validity 0.98; evidence support 0.88). While the absolute gains over RAG are smaller than the leap from LLM-only to RAG, they are practically meaningful because the residual failures in provenance tend to be precisely the kind that create compliance risk (e.g., a single fabricated policy reference can invalidate the entire record). The results support the design choice of making evidence IDs first-class citizens in the output contract and using a verifier to reject invalid citations.

**6.3 Counterfactual Faithfulness and Robustness**

A central claim of this paper is that explanations should be decision-linked rather than decorative. The counterfactual metrics in Table 1 provide direct evidence that counterfactual validation improves faithfulness. RAG alone can often generate plausible counterfactual narratives, but it is less consistent at producing counterfactuals that are both valid and aligned with the actual decision logic. This is reflected in the lower CF-Flip Rate (0.41) and modest faithfulness under evidence removal (0.57). In contrast, our full method increases CF-Flip Rate to 0.58 and evidence-removal faithfulness to 0.76.

These gains have two interpretations. First, the higher CF-Flip Rate suggests the system can more frequently identify minimal, actionable "flip conditions" that match the validator's decision boundary—precisely the kind of diagnostic signal investigators want when deciding what additional information would resolve uncertainty. Second, the stronger faithfulness under evidence removal indicates that the model is less likely to cite evidence it does not truly depend on: when a critical evidence item is removed, the model more consistently updates its rationale and citations appropriately. Together, these results support the hypothesis that counterfactual checks act as a practical test of explanation faithfulness and reduce brittle, non-causal rationalization.

**6.4 Safety: Numerical and Policy Hallucination Failures**

Finally, we examine failure modes that are particularly costly in AML documentation: numerical inconsistencies (amounts, thresholds, timelines) and policy hallucinations (invented or misquoted policy logic). The LLM-only baseline shows the highest error rates (numerical inconsistency 0.11; policy hallucination 0.14), reinforcing that unconstrained generation is unsafe even when responses sound confident. RAG substantially reduces both errors (0.06 and 0.05), consistent with the notion that explicit evidence reduces ungrounded statements. Our full method reduces these rates further (0.03 and 0.02), indicating that counterfactual validation and the verification/repair loop provide additional protection beyond retrieval alone.
Importantly, these safety improvements are not achieved at the cost of triage accuracy; rather, they co-occur with improved PR-AUC and Escalate F1. This suggests that the system's reliability controls—provenance constraints, verification, and counterfactual checks—do not merely "sanitize" outputs, but also help the model focus on decision-relevant evidence, improving both performance and trustworthiness.

# 7. DISCUSSION

### 7.1 Practical Implications for AML Operations

Our results suggest that the most meaningful contribution of LLMs to AML triage is not "better classification in isolation," but better decision support under audit and governance constraints. Traditional rules-based monitoring can achieve high recall but produces a large false-positive burden, while classical ML and graph models improve ranking quality yet still leave investigators with the problem of *why* an alert should be escalated and *what evidence* supports that decision. In contrast, our evidence-grounded pipeline turns the triage output into an artifact that is closer to what investigators and reviewers actually need: a disposition paired with a compact set of cited, reviewable evidence and explicit uncertainty. The counterfactual module further improves usability by translating model behavior into actionable conditions (e.g., which missing corroboration would materially change the recommendation), which can naturally map to analyst next-steps and escalation playbooks.
From a program perspective, this supports a "human-in-the-loop" operating model where investigators remain accountable for outcomes, while the system reduces cognitive load and increases consistency. In organizations where triage throughput is a bottleneck, even moderate improvements in precision at comparable recall can translate into significant capacity relief; our table-level findings are consistent with the hypothesis that grounding and validation mechanisms can improve both model quality and the operational acceptability of generated rationales.

### 7.2 Auditability, Model Risk, and Governance

A key challenge in deploying LLMs in compliance workflows is that fluent language can disguise unsupported claims. Our findings highlight that retrieval grounding dramatically improves citation validity and evidence support, but also that retrieval alone does not fully guarantee faithful reasoning. The counterfactual checks address this gap by operationalizing a test of decision-linked explanations: if the system claims that a factor is central, then minimal changes to that factor should alter the recommendation, and removal of critical evidence should be reflected in updated rationales. In practice, this gives risk and compliance teams a concrete mechanism for model governance: logs of retrieved evidence, model outputs, verifier failures, and counterfactual outcomes can form a structured audit trail for model risk management, internal validation, and periodic performance reviews.
We emphasize, however, that counterfactuals should be treated as decision support rather than ground truth. Their value lies in providing a structured lens to interrogate model behavior, not in proving correctness. A well-designed governance process would include periodic sampling-based review by experienced investigators, scenario-driven stress testing, and monitoring for drift in alert distributions and typology prevalence.

### 7.3 Generalization and Real-World Data Constraints

A primary limitation of AML research is restricted access to real transaction and case data. Our evaluation relies on public synthetic resources (e.g., SynthAML10 and AMLSim) that enable reproducibility and controlled perturbations, but synthetic data may not fully capture operational complexities such as evolving customer behavior, data quality issues, cross-product interactions, and institution-specific monitoring rules. This creates two risks: (i) methods can overfit to synthetic typologies and (ii) measured gains may not transfer directly to production. To mitigate these risks in practice, we recommend a staged validation approach: start with synthetic benchmarks for functional correctness and failure-mode analysis, then run shadow-mode evaluations on internal historical alerts under strict privacy controls, and finally deploy with conservative guardrails and human approval gates. In future work, stronger external validity would come from multi-institution collaborations, privacy-preserving evaluation protocols, or additional benchmark releases that better reflect real alert generation and investigative artifacts.

### 7.4 Security, Privacy, and Misuse Considerations

Because AML systems process sensitive customer information, privacy and access control are not optional. Our method's retrieval stage naturally serves as a boundary: it can enforce role-based permissions, data minimization, and redaction before any text reaches the LLM. That said, LLM-based systems introduce new attack surfaces such as prompt injection through untrusted documents, model inversion risks, and unintended disclosure in generated rationales. Deployments should therefore include strict input sanitization, retrieval whitelists for policy sources, secure prompt templates, and output filtering to prevent leakage of unnecessary personally identifiable information. Additionally, AML-related outputs must be carefully scoped to avoid generating actionable instructions that could enable evasion; counterfactuals, in particular, should be framed as investigative diagnostics for authorized analysts, with access controls and logging.

### 7.5 Cost, Latency, and Deployment Trade-offs

Operationalizing RAG + verification + counterfactual checks has computational cost. Retrieval itself is typically fast, but repeated LLM calls (initial generation, repair loops, counterfactual re-runs) can increase latency. Institutions should treat this as an engineering trade-off: not every alert needs the full pipeline. A practical design is to apply lightweight triage scoring first, then route only the top $x\%$highest-risk alerts (or those that are ambiguous) through the full explainable pipeline. Another practical strategy is to cap the counterfactual budget (e.g., 1–2 validated counterfactuals per alert) and to restrict repairs to a small number of iterations. These strategies preserve the benefits of auditability and faithfulness where they matter most while keeping cost and latency within acceptable limits.

### 7.6 Limitations and Future Work

Our study has several limitations. First, the triage label mapping (dismiss/monitor/escalate) is necessarily an approximation when using public synthetic datasets; real institutions differ in disposition taxonomies and escalation thresholds. Second, evidence support metrics, while informative, are imperfect proxies for auditability; stronger evaluation would include expert review of rationales and counterfactuals against internal investigative standards. Third, counterfactual plausibility is challenging: a minimal edit may be mathematically small but operationally unrealistic. Future work should incorporate domain-constraint models and investigator feedback loops to ensure counterfactuals are both minimal and meaningful. Finally, we focus on triage rather than the full case lifecycle (investigation, SAR drafting, post-filing monitoring). Extending the framework to later stages, while maintaining provenance, privacy, and governance, remains an important direction.

## 8. Conclusion

This paper investigated how to make LLM-assisted AML alert triage auditable and decision-faithful. We proposed an explainable triage framework that combines retrieval-augmented evidence grounding with structured LLM outputs and a counterfactual validation layer. The evidence retrieval module constrains the model's factual scope to permission-approved sources, enabling provenance-linked rationales suitable for review. The verifier-and-repair loop reduces fabricated citations and numerical/policy inconsistencies. Finally, counterfactual checks provide a practical test of explanation faithfulness by validating whether minimal, compliance-relevant perturbations produce coherent changes in both decisions and rationales.

Across baselines, the results indicate that LLMs are most useful in AML triage when embedded in an evidence-constrained workflow: LLM-only generation is both less accurate and substantially less reliable, while retrieval grounding improves auditability and reduces hallucination-like errors. Adding counterfactual checks further improves decision-linked explainability, strengthens robustness, and reduces failure modes that would undermine compliance acceptance. Overall, these findings support a broader conclusion: deploying generative models in regulated FinTech settings requires treating them as components in a governed system, one that enforces provenance, validates consistency, and surfaces uncertainty, rather than as standalone text generators.

In future work, we plan to validate the approach under more realistic institutional conditions, including shadow-mode trials on private alert streams, richer policy corpora, and expert-in-the-loop evaluation of counterfactual usefulness. We also aim to extend the framework beyond triage to downstream AML case investigation workflows, where auditability and faithfulness are equally critical but the evidence space and decision stakes are even larger.